# Subspace Learning for Feature Selection via Rank Revealing QR Factorization: Unsupervised and Hybrid Approaches with Non-negative Matrix Factorization and Evolutionary Algorithm

Amir Moslemi [a,b,*], Arash Ahmadian [c,**]

[a] Toronto Metropolitan University, Department of Physics Toronto, Ontario, Canada

[b] University of Toronto, Department of Medical Biophysics, Toronto, Ontario, Canada

[c] University of Toronto, Edward S. Rogers Sr. Department of Electrical and Computer Engineering, Toronto, Ontario, Canada

## Abstract

The selection of most informative and discriminative features from high-dimensional data has been noticed as an important topic in machine learning and data engineering. Using matrix factorization-based techniques such as nonnegative matrix factorization for feature selection has emerged as a hot topic in feature selection. The main goal of feature selection using matrix factorization is to extract a subspace which approximates the original space but in a lower dimension. In this study, rank revealing QR (RRQR) factorization, which is computationally cheaper than singular value decomposition (SVD), is leveraged in obtaining the most informative features as a novel unsupervised feature selection technique. This technique uses the permutation matrix of QR for feature selection which is a unique property to this factorization method. Moreover, QR factorization is embedded into non-negative matrix factorization (NMF) objective function as a new unsupervised feature selection method. Lastly, a hybrid feature selection algorithm is proposed by coupling RRQR, as a filter-based technique, and a Genetic algorithm as a wrapper-based technique. In this method, redundant features are removed using RRQR factorization and the most discriminative subset of features are selected using the Genetic algorithm. The proposed algorithm shows to be dependable and robust when compared against state-of-the-art feature selection algorithms in supervised, unsupervised, and semi-supervised settings. All methods are tested on seven available microarray datasets using KNN, SVM and C4.5 classifiers. In terms of evaluation metrics, the experimental results shows that the proposed method is comparable with the state-of-the-art feature selection.

\* All correspondence to  Amir.moslemi@ryerson.ca
\*\* Email address: arash.ahmadian@mail.utoronto.ca

# 1. Introduction

Challenges surrounding high-dimensional data has been strikingly noticed since the emergence of the internet and rapid development of information technology. Trace of curse of dimensionality is become bold in machine learning, pattern recognition, data mining, computer vision and recommender systems. Recently, dimensionality reduction techniques have been developed rapidly. Feature selection and feature extraction are the two main categories of dimensionality reduction. Feature extraction aims to map data to a lower dimension representation, whereas Feature selection is aimed to obtain the most discriminative features of high-dimensional data, which can be used to interpret medical data, industrial data, finance data, etc.

Feature selection in medical data can be important since the selected features can be used in disease diagnosis or optimizing treatments. To this end, a new line of research is devoted to bioinformatic and pattern recognition to deal with DNA microarray high dimensional datasets. These data types have significantly more features than sample points and thus need to be treated as an undetermined system [1,2]. Although thousands of genes from different samples are stored in a DNA micro array, only a small portion of the encoded information are related to disease [3]. A Gene selection problem in bioinformatic is equivalent to a feature selection task in machine learning.

In the context of disease diagnosis and progression analysis using feature selection, demographic, spirometry test and Computed Tomography (CT) features were collected and combined to predict chronic obstructive pulmonary disease (COPD) progression, exacerbation, and hospitalization. Then, extracting more prominent and discriminative features improved the accuracy of hospitalization prediction and discrimination between COPD and asthma [4-6].

Features without effect on output considered as irrelevant features, and redundant features are a mix of other features which cannot add more information. Including these irrelevant and redundant features increases the complexity of system which affects the performance of the learning algorithm and the computation time [7].

Feature selection is utilized for dimensionality reduction and to obtain the most informative subset of features by minimizing redundancy and maximizing the relevancy with the aim of increasing accuracy and decreasing computation time. Feature selection decreases task complexity and thus the probability of overfitting is decreased [8]. Peaking phenomenon states the error of a classifier for fixed data is decreased by adding features though error might be increased [9]. There are three strategies for feature selection; A) Filter, B) Wrapper, and C) Embedded, based techniques. In filter-based, the features are sorted and assessed based on some criteria by statistical and intrinsic characteristics of the dataset. In the filter method, there is no connection between classifier and dataset. Laplacian score [10], mutual information and MRMR are amongst the most well-known filter-based techniques [11-15]. In wrapper-based, there is a connection between features and the learning algorithm and thus the best subset of features can be achieved based on the output of the machine. Forward feature selection, sequential backward feature selection and floating search [16] and recursive SVM [17] are conventional feature selection techniques in machine learning.

Feature selection based on label information can be categorized to three types including supervised methods [18], semi-supervised methods [19], and unsupervised methods [20]. Information about relevancy of features and target can be obtained in a supervised setting making the goal of interaction between features and classes to find the most discriminative features. Under semi-supervised conditions part of data are labeled and information of both labeled and unlabeled data are extracted. Most of the semi-supervised techniques leverage a similarity matrix to select features by Laplacian graph (graph structure) construction [20]. There is no label information for the unsupervised case. Therefore, feature selection can be applied based on relevancy between features. This type of feature selection is most of the time considered as subspace learning with the aim to find a subspace of data can span original data [21]. The notations and list of abbreviations, which are frequently used in this study, are shown in Table 1 and Table 2.

Table 1. Notation used in the study.

| Notation | Representation |
| --- | --- |
| m | Number of instances |
| n | Number of features |
| $\sigma$ | Singular value of matrix |
| $I$ | Unit matrix |
| u | Lower case shows the vector |
| $A_{ij}$ | The (i, j)-th entry of matrix |
| $A^T$ | The transpose of matrix A |
| $Diag(A)$ | The vector of all diagonal elements of matrix A |
| $Tr(A)$ | The trace of square matrix $Tr(A) = \sum_{i=1}^{n} A_{ii}$ |
| $\det(A)$ | Determinant of matrix $A$ |
| $\|A\|_F$ | Frobenius norm of A. $\|A\|_F = \sqrt{Tr(A^T A)}$ |
| $A$ | Matrix $A \in R^{n \times m}$ |
| $\|u\|_p$ | Vector norm ($\ell_p$). $\|u\|_p = (\sum_{i=1}^{n} u_i^p)^{1/p}$ |
| $\|A\|_{q,p}$ | Mixed matrix Norm ($\ell_{q,p}$). $\|A\|_{q,p} = (\sum_{i=1}^{m} \|a^i\|_q^p)^{1/p}$ |

Recently, feature selection using matrix decomposition and subspace learning has gotten significant attention. Dealing with noisy information and redundant features are the main challenges for subspace learning. To circumvent this challenge, different regularization frameworks have been proposed to decrease the effect of noise and redundant information. Non-negative Matrix Factorization (NMF) combined with additional regularization terms is a well-known technique in this domain. Sparse regularization terms such as $\|.\|_{2,1}$, $\|.\|_{2,p}$ ($0 < p < 1$) and $\|.\|_{2,1-2}$ were added to increase the sparsity of solution [28,30,54,55]. However, information can be degraded or lost after projecting to subspace. To ameliorate this challenge, local structure learning regularization functions, which are based on graph Laplacian concept, were introduced.

These functions were added to preserve geometrical information of data in subspace with this assumption that if two samples were close, they would be close in the subspace. [31, 32, 53, 54].

Since feature selection is categorized as a combinatorial NP-hard problem, evolutionary computational techniques can be utilized to tackle this challenge. To this end, Genetic algorithms [38,46], particle swarm optimization (PSO) [39, 47], ant colony optimization [40, 48] and grey wolf optimization [41] have been applied on feature selection task to obtain the most informative features (genes). Cost function definition is one of the most important issues which requires significant consideration in feature selection using evolutionary techniques. The vast majority of feature selection cost function are constructed based on two well-known objectives: the number of features and classification error. Furthermore, many-objective feature selection methods have been proposed to enhance the classification accuracy [56]. This cost function encompassed the four objectives including the number of features, classification error, correlation between features and labels, and computational complexity of features.

Table 2. List of abbreviations, which are utilized throughout of this article.

| Complete Form | Abbreviation |
| --- | --- |
| Non-negative matrix factorization | NMF |
| Maximum projection and minimum redundancy feature | MPMR |
| Rescaled Linear Square Regression | RLSR |
| Dual regularized unsupervised feature selection based on matrix factorization and minimum redundancy | DR-FS-MFMR |
| ConCave–Convex Procedure | CCCP |
| Dual-graph sparse non-negative matrix factorization | DSNMF |
| Hesitant fuzzy set | HFS |
| Reduced row Echelon | RRE |
| Singular value decomposition | SVD |
| Particle swarm optimization | PSO |
| Genetic Algorithm | GA |
| Rank revealing QR | RRQR |
| Rank revealing QR- Genetic Algorithm | QR-GA |
| Principal component analysis | PCA |
| Non-negative matrix factorization QR | NMF-QR |

In all feature selection techniques, including NMF based and combinatorial methods, computational complexity is a considered heavily. For instance, the complexity of SVD and NMF are $O(mn^2)$ and $(k^2n + mn^2)$, respectively ($k$ is the number of selected features). To this end, a comparison between SVD-based principal component analysis (PCA) and QR-based PCA was done, and results showed that QR-based PCA is more efficient than SVD-based PCA in terms of computational complexity [57]. Additionally, these feature selection techniques have many hyper-parameters which must be tuned to achieve the best performance. Consequently, a matrix-based technique with less computational complexity and minimum hyper-parameters is the main motivation behind this work and to the best of our knowledge, it's the first to utilize QR factorization in feature selection. Additionally, followed by QR two hybrid techniques are proposed including; A) unsupervised NMF-QR feature selection, which is embedding QR in NMF, and B) A genetic algorithm is used, as an evolutionary algorithm, to combine with rank revealing QR factorization to construct a supervised hybrid feature selection by combination of filter policy and wrapper policy.

The contributions of this paper can be summarized as follows.

- For the first time rank revealing QR matrix factorization is proposed to find the most informative features of a high-dimensional data.
- Combination of NMF and QR as unsupervised feature selection.
- A Genetic algorithm is combined with rank revealing QR matrix factorization to obtain the optimum number of features and tune the hyperparameters of the classifier simultaneously with feature selection. Moreover, to prevent premature convergence to local optima, the quality of search space for the genetic algorithm is considerably enhanced by using Rank revealing QR matrix factorization as a filter-phase for hybrid feature selection.

The remainder of this paper is arranged as follows:

Several feature selections are surveyed in Section 2. Section 3 briefly reviews matrix decomposition. Section 3.1.2-3.1.3 belongs to RRQR and NMF-QR and 3.1.4 covers hybrid feature selection QR-GA. Results of the proposed method and its comparison with state-of-art methods are shown in Section 5. Finally, Section 7 and 8 draws the discussion and conclusion.

## 2. Related work

Matrix-based techniques, which are deeply rooted in linear algebra, have played an important role in feature selection task. Subspace learning plays a significant role in obtaining a low-dimensional representation of original data while preserving maximum information. To this end, non-negative matrix factorization (NMF) was formulated for feature selection by minimizing distance between the subspace and the original space of matrix [22]. In theory of NMF a matrix $A \in R^{m \times n}$ can be estimated by two non-negative matrices $U \in R^{m \times L}$ and $V \in R^{L \times n}$ such that $A \approx UV$. Therefore, NMF feature selection was defined as follows,

$$\min_{W,H} \frac{1}{2} \|A - AHW\|_F^2 + \frac{\alpha}{4} \|W^T W - I_k\|_F^2 \quad s.t \quad , W \geq 0, H \geq 0 \tag{1}$$

$\alpha$ is a positive penalty coefficient and this cost function was solved for fixed $H$ and $W$ in an iterative algorithm. $W$ is an indicator matrix (feature weight matrix) and the importance of each feature can be obtained using the Euclidean norm of $W$ rows. $H$ is representation matrix of data. A new term, which is to guarantee minimum redundancy, was added to NMF feature selection to enhance the performance of feature selection [23]. This method is called maximum projection and minimum redundancy feature (MPMR) and is mathematically expressed as following:

$$\min_{W,H} \frac{1}{2}\|A - AHW\|_F^2 + \alpha Tr(W^T A^T AW \mathbf{1}) + \frac{\beta}{4}\|W^T W - I_k\|_F^2 \ s.t\ , W \geq 0, H \geq 0 \quad (2)$$

where $\mathbf{1} \in R^{k \times k}$ whose elements are one and $Tr(W^T A^T AW\mathbf{1})$ plays the role of a correlation metric.

Structure learning regularization terms can be added to NMF feature selection to preserve information during projection to the subspace. These terms can be categorized to local and global learning regularization terms where graph Laplacian regularization and sparsity regularization are local and global, respectively. Local structure learning is added to preserve geometrical information. Therefore, graph Laplacian and Lasso norm were added to NMF feature selection to increase the performance [24].

$$\min_{W} \frac{1}{2}\|A - AW\|_F^2 + \frac{\beta}{2} Tr(W^T A^T SAW) + \alpha\|W\|_{2,1} \ s.t\ rank(W) = k \quad (3)$$

where $S$ is a structure learning regularization term, $\beta$ is a coefficient to adjust the weight of structure learning, $\alpha$ is a regularization parameter to control sparsity. $S$ can be calculated using the graph Laplacian such that $S = (I - Q)^T(I - Q)$. Where $Q$ is a symmetric affinity matrix of $A$. $\ell_{2,1}$ ($\|.\|_{2,1}$) norm (Lasso) was applied to preserve sparsity and robustness against outliers. The assumption underlying this Graph Laplacian based technique is that if two samples are close in the original data, the corresponding samples in lower-dimension space should be close. The main challenge of this structure learning regularization $S$ was that this term was not adaptive and updated in each iteration. This challenge was addressed by adaptive structure learning [53]. Adaptive structure learning ameliorates the challenge of noisy data as input by updating $S$ in each iteration. Additionally, $\|.\|_{2,1}^2$ was proposed to control outliers and noisy data [25].

Most of NMF-based feature selection concentrated on learning of feature weight matrix $W$ and no learning considered for representation matrix $H$. To circumvent this challenge, dual regularized unsupervised feature selection based on matrix factorization and minimum redundancy (DR-FS-MFMR) was proposed [60]. This study considered $R(W)$ and $R(H)$ as regularization terms for both the feature weight matrix and the representation matrix. Inner product regularization was leveraged to increase the sparsity of a solution and correlation between the rows of $W$ was modeled and added to objective function to minimize redundancy. Therefore, the objective function of DR-FS-MFMR can be written as follows:

$$\min_{W,H>0} \frac{1}{2}\|A - AHW\|_F^2 + \frac{\alpha}{2}(\mathbf{1}_k^T W^T A^T AW \mathbf{1}_k) + \frac{\beta}{2}\left(Tr(\mathbf{1}_{d \times d} WW^T) - Tr(WW^T)\right)$$
$$+ \frac{\mu}{2}\left(Tr(\mathbf{1}_{d \times d} HH^T) - Tr(HH^T)\right) \quad (4)$$

Where $\alpha$, $\beta$ and $\mu$ are trade-off coefficients between the reconstruction error term and regularization terms. In (4), the second, third, and fourth term represent the redundancy minimizer (correlation between rows of $W$), the regularization term on $W$, and the regularization term on $H$, respectively.

$L_{2,1}$ norm has been frequently used in feature selection to achieve sparse solutions and $\|.\|_2^2$ and $\|.\|_F^2$ are used to obtain the matrix subspace. However, $\|.\|_F^2$ suffers from outliers and noise reinforcing. To address this, $\ell_{2,1}$ norm was proposed to suppress the noise and outlier effects. To this end, feature selection using $\ell_{2,1}$ can be formulated as follows:

$$\min_{W} \|AW - Y\|_{2,1} + \lambda \|W\|_{2,1} \tag{5}$$

In another study, $L_{2,1}$ was utilized in least square problem for feature selection which is called semi supervised feature selection Rescaled Linear Square Regression (RLSR) [26]. In this technique, labeled matrix, which has label of each sample, is partitioned to labeled data and missed labeled data that is why it is called semi supervised feature selection.

$$\min_{W,\theta,b} \|A^T \theta W + \mathbf{1}b^T - Y\|_F + \lambda \|W\|_{2,1}$$

s.t

$$W, b, \theta > 0, \ \mathbf{1}^T \theta = 1, Y_U > 0, Y_U \mathbf{1} = \mathbf{1} \tag{6}$$

Where $Y \in R^{m \times c}$ is label matrix, since the problem was considered as semi-supervised, $Y$ was partitioned to $Y_l$ and $Y_u$ for labeled data and missed labeled data, respectively. The objective function (6) must be solved for $X$, $\theta$ and $b$. After convergence, $\theta$ showed the importance of features and it was sorted in descending order to take $k$ top features.

To increase sparsity, $\ell_{2,1}$ norm was replaced by $\ell_{2,p}$, where $p \in (0,1)$, in objective function of RLSR to have more sparse solution [27]. The global regularization term (sparse regularization term) plays a significant role in enhancing the sparsity of solution. Although the experimental results showed that the lowest classification error can be achieved by $\ell_{2,1/2}$ [28], this norm function is neither convex and nor Lipschitz continuous. Controlling the gradient can be a major challenge for non-continuous Lipschitz function. To ameliorate this challenge, $\ell_{2,1-2}$ matrix norm was proposed [29]. Although this norm function is nonconvex, it is Lipchitz continuous. Given the matrix $W$, its $\ell_{2,1-2} = \ell_{2,1} - \ell_{2,2}$ is given by

$$\|W\|_{2,1-2} = \|W\|_{2,1} - \|W\|_{2,2} \tag{7}$$

This study applied an iterative algorithm based on Concave–Convex Procedure (CCCP) to cope with nonconvex challenge by linearizing concave part as follows:

$$\min_{W} \|AW - Y\|_{2,1} + \lambda \|W\|_{2,1} - \lambda <W, \partial \|W\|_F> \tag{8}$$

Recently, $\|W\|_{2,1-2}$ as a non-convex (but Lipschitz continuous) constraint was combined with latent representation for feature selection. Although Laplacian graph regularization is frequently leveraged to preserve local geometric structure of the data and the feature space, it doesn't consider correlation between pseudo-labels. To combat this challenge, latent representation was considered to preserve more complete information between data space and feature space [61]. Based on latent representation theory, samples with similar latent representation have higher chance of getting affected by each compared to those with dissimilar latent representations.

The objective function of non-convex constraint and latent representation learning with Laplacian embedding can be expressed as follows:

$$\min_{W, H>0} \frac{1}{2} \|WA - Y\|_F^2 + \alpha \|W\|_{2,1-2} + \beta \|A' - YY^T\|_F^2 + \mu_1 Tr(W^T A^T L^A AW) + \mu_2 Tr(YL^Y Y^T) \quad (9)$$

Where $A'$ and $Y$ are the adjacency matrix and latent representation, respectively. $L^A$ is the Laplacian matrix of data and $L^Y$ is Laplacian matrix of the latent feature space.

As aforementioned, local regularization term is applied to preserve geometrical information during projection to subspace. Previous studies considered fixed similarity matrix which cannot be robust approach. To combat this challenge, adaptive structure learning was proposed [30]. In this technique, probabilistic neighbors-based manifold structure preservation was applied and the corresponding term in objective function is updated in each iteration. It should be noted that the local structure regularization term is defined only for data space. Consequently, this regularization term can only preserve the local manifold information of the data space but not that of the feature space. To this end, dual-graph sparse non-negative matrix factorization (DSNMF) was proposed [31]. Geometric information of both the data space and the feature space are preserved by the dual graph formulation. In DSNMF, the Laplacian matrix of the feature graph was calculated for both the data space and the feature space and corresponding terms are added to the objective function which can be mathematically formulated as follows:

$$\|A - WH^T\|_F^2 + \alpha \, Tr(W^T L^w W) + \beta \, Tr(H^T L^H H) + \lambda \|W\|_{2,1} \quad s.t \quad s.t \quad W \geq 0, \, H \geq 0 \quad (10)$$

Where $L^w$ and $L^H$ represent Laplacian matrix of data space and the feature space, respectively. To enhance the dual graph-based feature selection, hesitant fuzzy set (HFS) theory was added to this regression model of dual graph feature selection to enhance the performance of the algorithm [32].

Fundamental theories in linear algebra have also been explored in feature selection. Namely, feature selection using basis matrix construction was proposed [33]. In this technique, features are sorted based on the information gain of each feature. The first feature is selected from the sorted features (this feature has highest information gain) and rest of features are tested such that if a feature is spanned by the selected basis, it will be overridden. In another study, reduced row Echelon (RRE) was applied to extract independent columns (features) [34]. Since the column space of a matrix is independent of the order of columns, the columns are sorted based on information gain and thus RRE can be applied on the matrix to obtain independent column.

Therefore, linear independence can be considered as measurement of redundancy. In yet another study, singular value decomposition (SVD) and NMF was combined for feature selection [35]. In this technique, the subspace of matrix $A$ is obtained using $SVD(A) = U\Sigma V^T$, and column space of matrix is extracted by considering first r-column of $U$ which is denoted by $U_r$. To apply NMF feature selection, it can be considered that $AW = U_r$ and $H = \Sigma V^T$. Therefore, the objective function of this feature selection can be expressed based on NMF feature selection such that $V^T \approx (AW\Sigma)^t A$, where t is Moore-Penrose pseudoinverse.

$$\min_{W} \frac{1}{2}\|A - AWH\|_F^2 + \frac{\beta}{4}\|W^T W - I_k\|_F^2 - \frac{1}{2}\|Wg\|_F^2 \quad s.t \quad W \geq 0 \qquad (11)$$

Where $g$ is a vector, whose elements are information gain of each feature and this term added to objective function to achieve maximum relevancy. In the context of applications of matrix analysis to feature selection, a feature selection technique based on perturbation theory was proposed in [36]. In this technique, the least square problem was solved for matrix $A$ and perturbed matrix $A + E$ was calcuted, where $E$ is a random matrix with norm-2 equal to minimum singular value ($\|E\|_2 = \sigma_{min}$) of matrix A. The solutions of least square problem of matrix A and perturbed matrix $A+E$ are subtracted to calculate $\Delta x = \|Ax - b\| - \|(A + E)x' - b\|$. $\Delta x = x - x'$ was calculated for each feature. Consequently, the features with similar $\Delta x$ are grouped through clustering using techniques such as K-means clustering. Features in each cluster are ranked based on entropy and first rank of each cluster was considered as the selected feature.

Gram-Schmidt is a technique in linear algebra to construct a set of orthonormal vectors. This theory was utilized for feature selection and Random Projection and Gram-Schmidt Orthogonalization (RP-GSO) from the word co-occurrence matrix was proposed [37].

High computational cost and high number of regularization parameter are gaps studies which this work aims to address.

## 3. Proposed method

The proposed method directly utilizes a characteristic of QR matrix factorization. SVD is the most popular method to decompose the matrix to obtain a subspace of the matrix which can span the matrix. Although column pivoting QR decomposition is considered as a cheaper alternative for SVD, it may fail in select cases as showed in Example 3.1 by Kahan [58]. To tackle this challenge strong rank revealing QR factorization was proposed, which is the most promising alternative for SVD without any failing [43,59]. In this study, we utilized strong rank revealing QR factorization for feature selection aim.

The optimum number of selected features plays an essential role in enhancing the performance of the learning algorithm. To this end, we propose a two-phase feature selection which is a combination of filter and wrapper techniques. In this hybrid technique, rank revealing QR factorization is applied to extract linearly independent features as an unsupervised filter-phase, and thus a genetic algorithm is applied to obtain optimum subset of features as wrapper phase.

## 3.1 Methodology

### 3.1.1 Subspace Learning

Feature selection can be seen as a problem which looks for a subset of features that can be an approximate of the whole features. For better clarification, a matrix $A \in R^{m \times n}$ has $n$- features $\{f_1, \ldots, f_n\}$ and matrix $A_J \in R^{m \times k}$ has k-feature such that $A_J \subset A$. Meaning that $A_J$ is a sub-matrix of $A$ and for it to be the best approximation of $A$. To this end, $A_J$ must be sufficiently close to $A$ and it can be formulized as follows:

$$\underset{J}{\operatorname{argmin}} \, distance \, (span(A), span(A_J)) \tag{12}$$

Different metrics have been introduced to measure distance between $A$ and $A_J$. In this work, Rank Revealing QR factorization is leveraged to obtain $A_J$.

### 3.1.2 Rank-Revealing QR factorization

Rank revealing QR (RRQR) factorization of real matrix $A$ is a factorization of this matrix to the the product of two matrices. RRQR has the form

$$A\Pi = QR \tag{13}$$

Where $\Pi$, $Q$ and $R$ are the permutation matrix, the orthonormal matrix, and the upper triangle matrix, respectively. For a matrix $A \in R^{m \times n}$ and approximation rank $k$, the matrix $R$ can be represents as following

$$A\Pi = Q \begin{bmatrix} R_{11} & R_{12} \\ 0 & R_{22} \end{bmatrix} \tag{14}$$

and we have $R_{11} \in R^{k \times k}$ (with nonnegative diagonal elements), $R_{12} \in R^{k \times n-k}$, $R_{11} \in R^{n-k \times n-k}$ and $Q \in R^{m \times n}$.

RRQR factorization can be satisfied if and only if following conditions are satisfied.

$$\sigma_{min}(R_{11}) \geq \frac{\sigma_k(A)}{p(k,n)}, \sigma_{min}(R_{11}) \geq \sigma_{k+1}(A) \times p(k,n) \tag{15}$$

Where $p(k,n)$ is a function bounded by a low-degree polynomial in "$k$" and "$n$" and $\sigma$ is a singular value. Although RRQR has worked well in many cases, there are instances where it fails to generate a permutation matrix (Example 3.1 by Kahan). [42]. On the other hand, RRQR cannot be a stable algorithm due to a significantly large condition number of matrix $R_{11}$ ($cond(A) = \frac{\sigma_{max}(A)}{\sigma_{min}(A)}$). Aiming to tackle these defects, an efficient algorithm with the underlying goal of maximizing the determinant of $R_{11}$ was proposed which can be utilized without failing [43].

#### 3.1.2.1 Permutation matrix computation

For a given rank $k$, the RRQR can be expressed as following:

$$A\Pi = [Q_1 \; Q_2] \begin{bmatrix} R_{11} & R_{12} \\ 0 & R_{22} \end{bmatrix} \tag{16}$$

Where $Q_1$ and $Q_2$ are the first $k$-column of $Q$ and "$k+1:n$" columns of $Q$, respectively. Therefore, truncated matrix of $A$ of can be obtained by the following optimization problem:

$$min \; \|A - Q_1[R_{11} \; R_{12}]\|_2 \tag{17}$$

Therefore,

$$\|A - Q_1[R_{11} \; R_{12}]\|_2 = \|Q_2 R_{22}\|_2 = \|R_{22}\|_2 = \sigma_{max}(R_{22}) \tag{18}$$

It can clearly be observed that $\sigma_{max}(R_{22})$ must be sufficiently small and negligible, and this is the first condition to compute permutation matrix with aim of obtaining RRQR. Second condition is that $\sigma_{min}(R_{11})$ must be sufficiently large and $R_{11}^{-1} R_{22}$ is bounded. Consequently, the main aim of RRQR is to find a permutation matrix such that first and second conditions are satisfied.

### 3.1.2.2 Feature Selection Using RRQR

Although SVD and RRQR provide low rank representations of the original matrix, only RRQR can determine the independent columns using its permutation matrix. For a given rank $k$, SVD of matrix A is

$$SVD(A) = [u_1, \dots, u_k, u_{k+1}, \dots] \Sigma V^T \tag{19}$$

Where $u_i$, $\Sigma$ and $V$ are unit vector, diagonal matrix whose elements are singular values, and orthonormal matrix, respectively (U=$[u_1, \dots, u_k, u_{k+1}, \dots]$ is an orthonormal matrix). $U_k = [u_1, \dots, u_k]$ is low-dimensional representation of $A$ such that it can span $A$. The subtle point should be noted is that the independent columns of matrix $A$ cannot be directly obtained by $SVD$. However, the permutation matrix of RRQR can determine the independent columns of $A$.

The RRQR can be expressed as follows:

$$A[\Pi_1 \Pi_2] = [Q_1 \; Q_2] \begin{bmatrix} R_{11} & R_{12} \\ 0 & R_{22} \end{bmatrix} \tag{20}$$

Where $\Pi_1$ is a permutation matrix which shows the independent columns of matrix A.

Therefore,

$$A\Pi_1 = Q_1[R_{11} \; R_{12}] \tag{21}$$

$$A\Pi_2 = Q_2 R_{22} \tag{22}$$

Based on the first condition of permutation matrix computation in RRQR, $\|A\Pi_2\|_2$ must be sufficiently small and negligible, thus the most of information (energy) are embedded in $A\Pi_1$.

$\omega$ and $\gamma$ are two new terms which are defined as follows:

$$\omega(A) = (\omega_1(A), \ldots, \omega_n(A)) \tag{23}$$

where $\omega_i(A) = \left\| A^{-1}_{i,:} \right\|_2$ and $A^{-1}_{i,:}$ indicates $i^{th}$ row of $A^{-1}$,

and

$$\gamma(A) = (\gamma_1(A), \ldots, \gamma_n(A)) \tag{24}$$

where $\gamma_j(A) = \left\| A_{:,j} \right\|_2$ and $A_{:,j}$ indicates $j^{th}$ column of $A$.

Algorithm 1 reveals the procedures of strong RRQR, and details of this algorithm are discussed in section 3.1.2.3.

**Algorithm 1.** Strong RRQR factorization

---
**Input:** $A \in R^{m \times n}, \Pi = I\ (I \in R^{n \times n}), k = \#\ selected\ features, f$
**Repeat**
1. Find $i, j$ such that $(R_{11}^{-1} R_{12})^2_{i,j} + (\frac{\gamma_j(R_{22})}{\omega_i(R_{11})})^2 > f^2$
2. $\begin{bmatrix} R_{11} & R_{12} \\ 0 & R_{22} \end{bmatrix} \leftarrow \begin{bmatrix} R_{11} & R_{12} \\ 0 & R_{22} \end{bmatrix} \times \Pi_{i,j+k}$
3. $\Pi \leftarrow \Pi \times \Pi_{i,j+k}$

**Until** no $i, j$ can find to satisfy $(R_{11}^{-1} R_{12})^2_{i,j} + (\frac{\gamma_j(R_{22})}{\omega_i(R_{11})})^2 > f^2$

---

Permutation matrix $\Pi_{i,j+k}$ interchanges the $i^{th}$ and $j^{th}$ columns of a matrix.

Where parameter $f$ is a constant scalar that bounds the entries of the calculated $(R_{11}^{-1} R_{12})$. The proposed algorithm for RRQR feature selection is given below.

**Algorithm 2.** Feature selection based on RRQR factorization.

---
**Input:** Data matrix $A \in R^{m \times n}, k = \#\ selected\ features$
1. Apply Algorithm 1 on matrix $A$
2. $A\Pi_1 = Q_1[R_{11}\ R_{12}]$

**Output:** Consider the first $k$-index of $\Pi$ (Index of $\Pi_1$) as independent column of matrix A.

---

If $k$ is considered equal to the rank of matrix, all of columns which are independent of each other are chosen. Otherwise, possibly some column features are lost.

For feature selection using RRQR, the time complexity is $O(mnk)$ where $k$ is the number of selected features.

### 3.1.2.3 Existence of Strong RRQR

As aforementioned, three conditions must be satisfied to have a strong RRQR: 1) sufficiently large singular values for $R_{11}$, 2) sufficiently small singular values for $R_{22}$, and 3) bounded elements for $R_{11}^{-1}R_{12}$. By considering these three conditions and having following expression for $\det(R_{11})$,

$$\det(R_{11}) = \prod_{i=1}^{k} \sigma_i(R_{11}) = \sqrt{\det(A^T A)} / \prod_{j=1}^{n-k} \sigma_j(R_{22}) \tag{25}$$

a strong RRQR can be resulted in a large $\det(R_{11})$. Consequently, Algorithm 1 was constructed to maximize $\det(R_{11})$ via columns interchanging (permutation matrix). Then, the columns of matrix $A$ would be interchanged until no increase could be occurred in $\det(R_{11})$. To prove Algorithm 1 let consider Lemma 1.

**Lemma 1:** Let

$$R = \begin{bmatrix} R_{11} & R_{12} \\ 0 & R_{22} \end{bmatrix} \text{ and } \Omega_k(R\Pi_{i,j+k}) = \begin{bmatrix} R'_{11} & R'_{12} \\ 0 & R'_{22} \end{bmatrix} \tag{26}$$

Where all diagonal elements of $R_{11}$ are positive. Then

$$\frac{\det(R'_{11})}{\det(R_{11})} = \sqrt{(R_{11}^{-1}R_{12})_{i,j}^2 + \frac{\gamma_j(R_{22})}{\omega_i(R_{11})}} \tag{27}$$

**Proof.**

First, consider $i < k$ or $i < j$ and let $R_{11}\Pi_{i,k} = \hat{Q}\hat{R}_{11}$ (QR factorization of $R_{11}\Pi_{i,k}$), $\hat{R}_{12} = \hat{Q}^T R_{12}\Pi_{1,j}$, $\hat{R}_{22} = R_{22}\Pi_{1,j}$ and $\hat{\Pi} = diag(\Pi_{i,k}, \Pi_{1,j})$. Then the QR factorization for $R\hat{\Pi}$ can be expressed as following:

$$R\hat{\Pi} \equiv \begin{bmatrix} R_{11}\Pi_{i,k} & R_{12}\Pi_{1,j} \\ 0 & R_{22}\Pi_{1,j} \end{bmatrix} = \begin{bmatrix} \hat{Q} & 0 \\ 0 & I_{m-k} \end{bmatrix} \begin{bmatrix} \hat{R}_{11} & \hat{R}_{12} \\ 0 & \hat{R}_{22} \end{bmatrix} \tag{28}$$

It can be resulted that $\det(R_{11}) = \det(\hat{R}_{11})$ since both $R_{11}$ and $\hat{R}_{11}$ have positive diagonal elements. From (25),

$$\hat{R}_{11}^{-1}\hat{R}_{12} = \Pi_{i,k}^T R_{11}^{-1} R_{12}\Pi_{1,j} \tag{29}$$

Then,

$$(R_{11}^{-1}R_{12})_{i,j} = (\hat{R}_{11}^{-1}\hat{R}_{12})_{i,j} \tag{30}$$

Since we have

$$\hat{R}_{11}^{-1} = \Pi_{i,k}^T R_{11}^{-1} R_{12}\hat{Q} \tag{31}$$

and $\ell_2$ norms of the matrix rows would not be changed by multiplication to orthonormal matrix, we have

$$\omega_i(R_{11}) = \omega_k(\hat{R}_{11}) \tag{32}$$

and

$$\gamma_j(R_{22}) = \gamma_1(\hat{R}_{22}) \tag{33}$$

Therefore, it only needs to consider $i = k$ and $j = 1$ as a special case.

By considering following partition

$$\Omega_{k+1}(R) = \begin{pmatrix} R_{11_{k-1}} & b_1 & b_2 & R_{22} \\ 0 & z_1 & z_3 & c_1^T \\ 0 & 0 & z_2 & c_2^T \\ 0 & 0 & 0 & C_{k+1} \end{pmatrix}$$

Then $\omega_i(R_{11}) = z_1$, $\gamma_j(R_{22}) = z_2$ and $(R_{11}^{-1}R_{12})_{i,j} = \frac{z_3}{z_1}$. Determinant of $R_{11}$ and $R'_{11}$ can be calculated as follows;

$$\det(R_{11}) = \det(R_{11_{k-1}}) z_1 \tag{34}$$

and

$$\det(R'_{11}) = \det(R_{11}\Pi_{i,j+k}) = \det(R_{11_{k-1}}) \sqrt{z_3^2 + z_2^2} \tag{35}$$

, so that

$$\frac{\det(R'_{11})}{\det(R_{11})} = \sqrt{(\frac{z_3}{z_1})^2 + (\frac{z_2}{z_1})^2} = \sqrt{(R_{11}^{-1}R_{12})_{i,j}^2 + (\frac{\gamma_j(R_{22})}{\omega_i(R_{11})})^2} \tag{36}$$

The aim of algorithm 1 is to find $i \ and \ j$ such that the maximized of (36) can be achieved.

### 3.1.3 NMF-QR

In this technique QR decomposition was embedded in NMF feature selection to obtain the most discriminative features. Fundamentally, the advantage of this technique is that the computation of permutation matrix is not necessary and then economic RRQR can be applied.

**Proposition.** If matrix $A$ has $rank(A) = k$, first $k$-columns of $Q$ form an orthonormal basis for $span(A)$ [62].

In QR, $Q_1$ is a good approximation of the original data matrix $A_J$, can span $A$. Likewise, the subspace of matrix $A$ can be extracted using the following formula:

$$\|A - Q_1[R_{11}R_{12}]\|_F^2 = \|A - A_J[R_{11}R_{12}]\|_F^2 \ s.t \ |J| = k \tag{37}$$

Therefore, by considering NMF objective function (1), it can be found $AW = Q_1$ and $H = R_{11}$. In this technique we want to find the best approximation of $Q_1$. Then, $R_{11} = (AW)^{-1}A$ can be considered as a rule for updating $R_{11}$ with respect to the new $Q_1 = A_J$.

Graph Laplacian as local learning regularization can be added to feature selection to preserve data geometry information during projection to subspace [63]. This technique is constructed based on the assumption that if two samples $a_i$ and $a_j$ are close in original space of data, then the corresponding samples $b_i$ and $b_j$ in the sub-space should be close. $A \in R^{n \times m}$ as original data is projected by mapping function $F$ to subspace $A' \in R^{n \times m'}$ such that $m >> m'$. Therefore, $\|F\|_M^2 = \int \|\nabla_M F\|^2$ is applied to measure smoothness of "$F$" along the geodesic of data with submanifold $M \subset R^m$ such that $\nabla_M F$ is gradient of $F$ with submanifold $M$. Since $M$ is not known, therefore $\|F\|_M^2$ cannot be obtained. To tackle this challenge $\|F\|_M^2$ can be discretely approximated as following [64,65]:

$$\|F\|_M^2 = \frac{1}{2}\sum_{i=1}^n \sum_{j=1}^n \|b_i - b_j\|_2^2 \Gamma_{ij} = \frac{1}{2}\sum_{i=1}^n b_i b_i^T \sum_{j=1}^n \Gamma_{ij} + \frac{1}{2}\sum_{j=1}^n b_j b_j^T \sum_{i=1}^n \Gamma_{ij} - \sum_{i=1}^n \sum_{j=1}^n b_i b_j^T \Gamma_{ij} \qquad (38)$$

Where $b$ and $\Gamma \in R^{n \times n}$ are sample in subspace (a row vector) and symmetric affinity matrix, respectively. $\sum_{j=1}^n \Gamma_{ij}$ is equal to diagonal matrix whose diagonal elements are equal to sum of affinity matrix $\Gamma$. Therefore, $D_{ii} = \sum_{j=1}^n \Gamma_{ij}$. the formula (38) can be written as follows:

$$\|F\|_M^2 = \frac{1}{2}\sum_{i=1}^n b_i b_i^T D_{ii} - \sum_{i=1}^n \sum_{j=1}^n b_i b_j^T \Gamma_{ij} = Tr(B^T(D-\Gamma)B) = Tr(B^T S B) \qquad (39)$$

Where $B = \sum_{i=1}^n b_i$ is matrix of samples in subspace and $S = D - \Gamma$ is called graph Laplacian of matrix $A$. $B$ can be substituted by $AW$ which shows the projection of data in subspace. Then locally preserving regularization term is $Tr(W^T A^T S A W)$ and Kernel $\Gamma$ is defined as follows:

$$\Gamma_{ij} = \begin{cases} \exp\left(-\frac{\|a_i - a_j\|_2^2}{\sigma}\right) & if \ a_j \in N_k(a_i) \ or \ a_i \in N_k(a_j) \\ 0 & \end{cases} \qquad (40)$$

Where $N_k(a_i)$ is the set of K-nearest neighbours of $a_i$ and $a_i \notin N_k(a_i)$ and $\sigma$ is the kernel parameter. Then, feature selection with a structure preservation term can be rewritten as follows:

$$\min_W \frac{1}{2}\|A - AWH\|_F^2 + \frac{\alpha}{2} Tr(W^T A^T S A W) \quad s.t \quad W^T W = I, \ W > 0 \qquad (41)$$

Sparse regularization term should be embedded to (41) to preserve global information. Although $\ell_{2,1}$ is frequently used to generate a sparse solution, $\ell_{2,1/2}$ was considered as a global learning regularization term in this study, since results showed that it performs better than $\ell_{2,1}$ [30]. To provide a better intuition for sparsity regularization terms, contour maps of $\ell_{2,1}$, $\ell_{2,1/2}$ and $\ell_{2,2}$ matrix norms are plotted and shown in Figure 1.

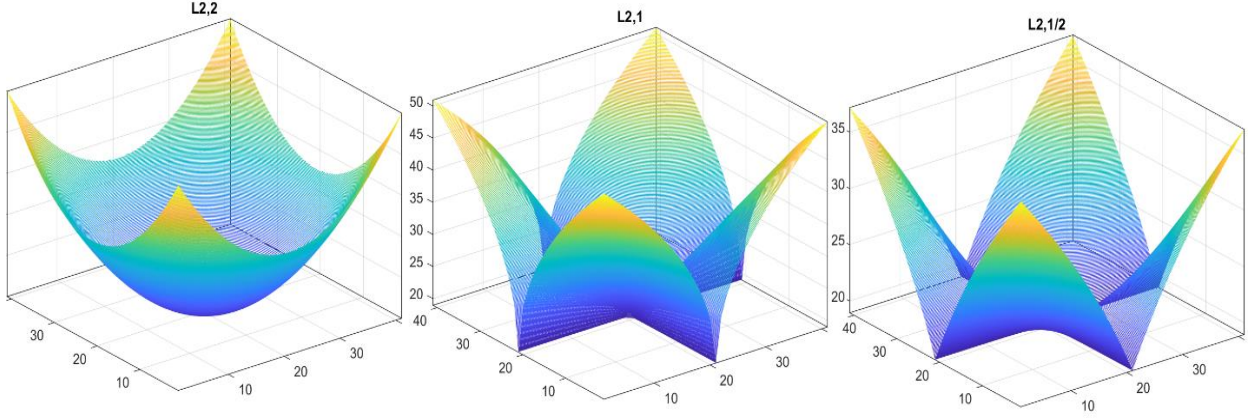

**Figure 1.** The contour plots of $\ell_{2,1}$, $\ell_{2,1/2}$ and $\ell_{2,2}$ matrix norms which are utilized as regularization term in optimization process.

Consequently, $\|W\|_{2,1/2}^{1/2}$ is added to (41) to formulize the NMF-QR:

$$\min_{W} \frac{1}{2}\|A - AWH\|_F^2 + \frac{\alpha}{2}Tr(W^T A^T S A W) + \gamma\|W\|_{2,1/2}^{1/2} \quad s.t \quad W^T W = I,\ W > 0 \quad (42)$$

$\|W\|_{2,\frac{1}{2}}^{\frac{1}{2}}$ can be replaced by $Tr(W^T \Phi W)$ such that $\Phi$ is a diagonal matrix and defined as:

$$\Phi = diag\left(\frac{1}{4\|w^i + \varepsilon\|_2^{3/2}}\right) \quad (43)$$

Where $\varepsilon$ is numerical stability constant and $w^i$ represents the $i^{th} -$ row of matrix $W$. It should be noted that $\ell_{2,\frac{1}{2}}$ is not a real norm since it cannot satisfy the triangle inequality property.

To solve (42), the following Lagrange function method is devised.

$$\mathcal{L}(W, \Lambda) = \frac{1}{2}\|A - AWH\|_F^2 + \frac{\alpha}{2}Tr(W^T A^T S A W) + \frac{\gamma}{2}Tr(W^T \Phi W) + \frac{\beta}{4}\|W^T W - I\|_F^2 +$$

$$Tr(\Lambda W^T) \quad (44)$$

Where $\alpha > 0$, $\gamma > 0$ and $\beta > 0$ are coefficient parameters to have a trade-off between reconstruction error term and regularization terms.

The solution can be obtained by taking derivative of $\mathcal{L}$ with respect to $W$.

$$\frac{\partial \mathcal{L}}{\partial W} = \frac{1}{2}(-2A^T A H^T + 2A^T A W H H^T) + \alpha(A^T S A W) + \gamma \Phi W + \beta(W W^T W - W) + \Lambda$$

By applying Karush–Kuhn–Tucker (KKT) conditions and setting $\frac{\partial \mathcal{L}}{\partial W} = 0$, we have:

$$\sum \Lambda_{ij} W_{ij} = 0 \tag{45}$$

Eventually, update rule of $W$ can be derived as follows:

$$W_{ij} \leftarrow W_{ij} \frac{A^T A H^T + \beta W}{A^T A W H H^T + \alpha(A^T S A W) + \gamma \Phi W + \beta(W W^T W)} \tag{46}$$

After a given number of iterations, Euclidean norms are calculated for each row of $W$ that is associated with a feature, and the rows are sorted in descending order. The algorithm of NMF-QR is shown in Algorithm 3.

**Algorithm 3.** NMF-QR for feature selection.

---
**Input**: Feature matrix $A \in R^{m \times n}$, , and parameters $\beta, \alpha, \gamma$
**Output**: Top Features calculate
1. Initialize $W$
2. $r = rank(A), QR(A) = Q_1[R_{11} R_{12}]$
3. Compute structure preserving regularizer $S$
4. **Repeat**
5. $H = R_{11} R_{12}$
6. $W = W D^{-1}$ such that $D = (diag(W W^T))^{1/2}$  # $W$ normalization
7. Calculate $\Phi = diag(\frac{1}{4\|w^i + \varepsilon\|_2^{3/2}})$
8. Update $W_{ij} \leftarrow W_{ij} \frac{A^T A H^T + \beta W}{A^T A W H H^T + \alpha(A^T S A W) + \gamma \Phi W + \beta(W W^T W)}$
9. Update $R_{11} R_{12} = (AW)^{-1} A$
10. **Until** Convergence criterion has been satisfied (Maximum number of iterations)
11. Sort rows of $W$ based on $\|w^i\|_2$
12. **return** Select the top features
---

In terms of hyper-parameters, Parameter $\beta > 0$ is a penalty term to make a trade-off between error of reconstruction and the closeness of the indicator matrix to orthogonality. Different $\beta$ values were experimented with, and results showed that a higher value of $\beta$, guarantees more orthogonality.

Computational cost for QR, structure preservation, and $W$ updating are $O(knm)$, $O(n(nm))$ and $O(Tcmn)$, respectively. Where $n, m, c, T, k$ are the number of samples, number of features, number of classes, number of iterations, and rank, respectively. Therefore, the total complexity of algorithm is $O(knm + n^2 m + Tcmn)$.

### 3.1.4 Hybrid Feature Selection

In this technique, RRQR is applied to extract full rank of features matrix such that $k$ is considered as the rank of matrix. Consequently, a genetic algorithm (GA) is applied to find optimum subset of features. This hybrid feature selection is called QR-GA.

It should be noted that when we use the optimum term in this study, it means best recorded performance. Since when we are using GA for a combinatorial NP-hard problem, there is no guarantee that the solution is optimal

The fitness function of feature selection using GA is defined as follows:

$$fitness\ function = \omega(Error) + (1-\omega)\left(\frac{Subset\ of\ features}{Total\ features}\right) \quad (47)$$

Where $\omega$ is a predefined parameter to provide a conciliation between error of classifier and the number of selected features, parameter $0 < \omega < 1$ is a regularization parameter to make a trade-off between accuracy of classifier and the number of selected features with the aim of preserving highest performance of classifier and least selected features. $\frac{Subset\ of\ features}{Total\ features} \leq 1$ is a ratio of selected features and the main aim is to reduce the dimension while maintaining classifier accuracy as much as possible.

An important point should be noted that the hyperparameter of a classifier can be tunned during feature selection. Consider a support vector machine (SVM) which has two hyperparameters that are effective for performance of the decomposition. These hyperparameters can be optimized simultaneously with the feature selection task. Therefore, two aims, optimum subset of features and tunned hyperparameters, can be achieved in the phase-2 of hybrid feature selection.

A chromosome of GA, which is carrying a solution, can be portioned such that each part belongs to specific hyperparameters or selected features. for example, a chromosome is portioned to three parts such that first, second and third parts represent the first hyperparameter, second hyperparameter, and the subset of selected features, respectively. Partition chromosome for SVM classifier is shown in Figure 2. C and gamma are two hyperparameters of SVM where C controls a trade of between complexity and non-separable samples and gamma is radius of RBF kernel.

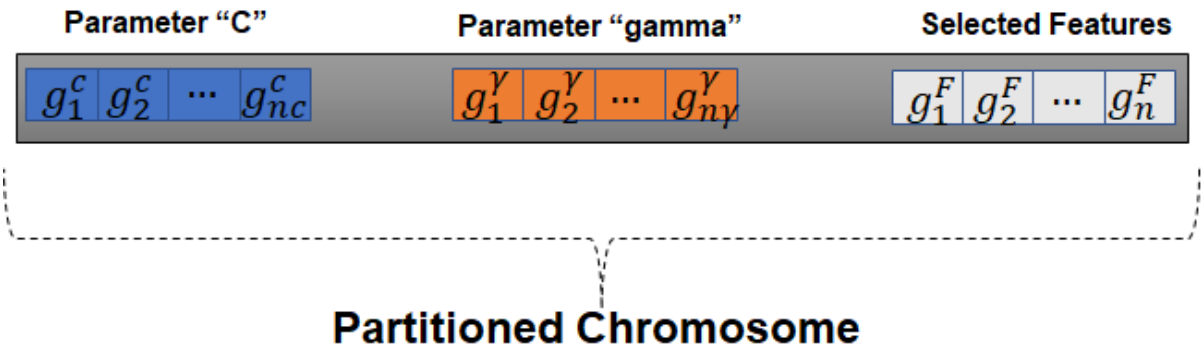

**Figure 2.** A chromosome is partitioned to three parts and $g$ is gene.

It should be noted is that genes can take only two values "0" and "1". Accordingly, "1" shows the selected features for feature selection part, whereas to parameters "C" and gamma, these binary values are meaningless. To circumvent this challenge, these binary values are transformed from genotype to phenotype.

$$Phenotype\ of\ parameter = min_P + \frac{max_P - min_P}{2^l - 1} \times \rho \qquad (48)$$

Where $min_P$, $max_P$, $l$ and $\rho$ are the minimum and maximum values of parameter that it can take, , the number of bits that belong to the parameter, and decimal value of the parameter, respectively.

The flowchart of hybrid feature selection is shown in Figure 3.

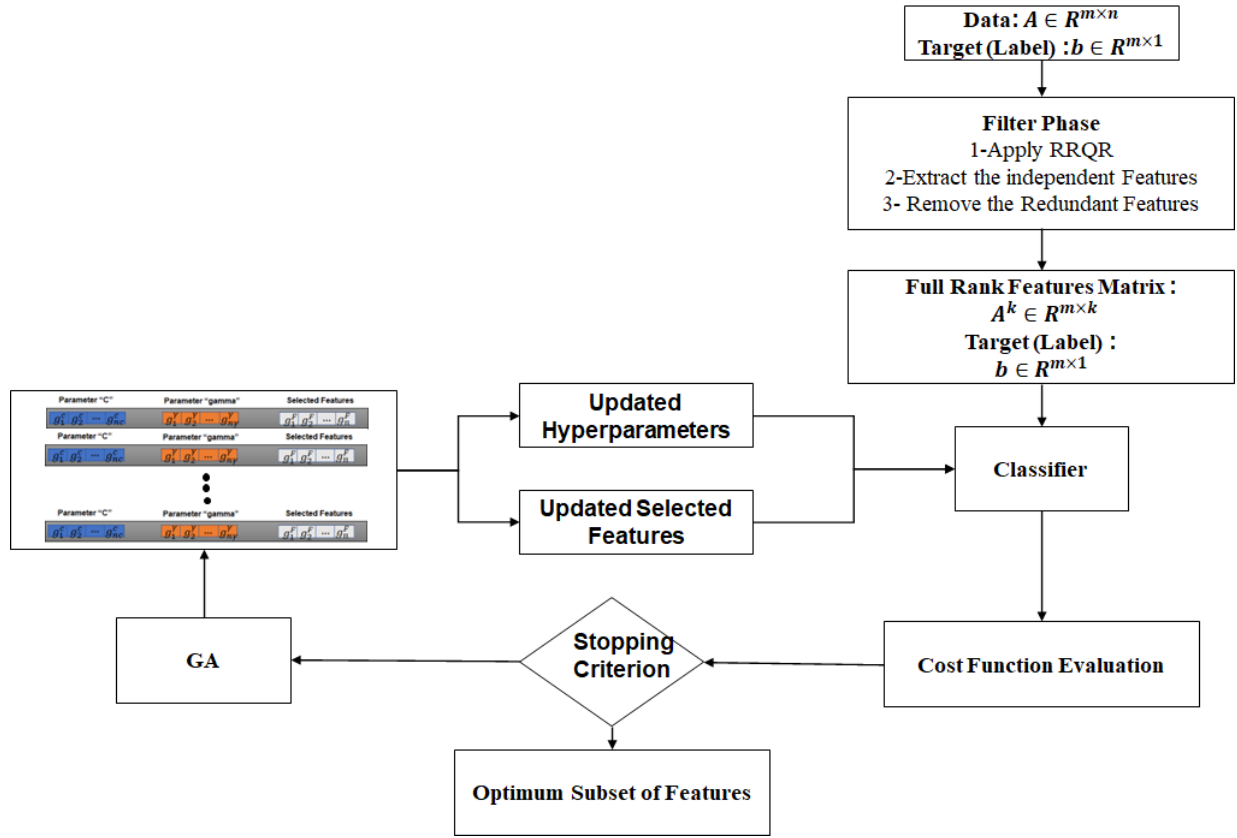

**Figure 3.** This figure shows the two-phase hybrid feature selection flowchart.

## 4. Experiments

In this section, the performance of the proposed algorithms, RRQR and QR-GA, are assessed by comparing with state-of-the-art feature selection techniques including: NMF feature selection [22], MPMR [23], NMF with structure learning regularization and sparse regularization (NMFSLS) [24], Rref [34], SVD [35], Perturbation [36], MRMR [44] and ReliefF [45]. Finally, top-10 and top-50 features are reported. Results were not directly adopted from related research and all these techniques were simulated.

### 4.1 Dataset

Eight microarray datasets are used for experiments, which are frequent common benchmark for the performance of feature selection techniques. Table 3 indicates the dataset, number of samples, number of features and number of classed. Details of the dataset can be found in [53].

**Table 3.** Summary description of the dataset which were used for experiments.

| Dataset | #Samples | #Features | #Classes | References |
|---------|----------|-----------|----------|------------|
| Ovarian | 253 | 15154 | 2 | [49] |
| Colon | 62 | 2000 | 2 | [50] |
| Breast | 97 | 24481 | 2 | [51] |
| Prostate | 102 | 6033 | 2 | [52] |
| SMK | 187 | 19993 | 2 | [53] |
| Lung | 203 | 3312 | 2 | [54] |
| GLI | 50 | 22283 | 2 | [53] |
| Leukemia | 72 | 7070 | 2 | [56] |

### 4.2 Cross-Validation

Since datasets, which are considered in Table 3, are not portioned to train-test sets, using cross validation techniques is indispensable. To this end, DOB-SVC was suggested to apply on microarray datasets to guarantee the distribution of each class in each fold [57-58]. Consequently, 5-fold DOB-SVC cross-validation technique is leveraged in our experiments.

### 5. Performance Measures

### 5.1 Evaluation Metrics

Since some datasets can be imbalanced, different evaluation metrics are required. To this end, accuracy, sensitivity, specificity, F1-score, and G-mean are considered to assess the proposed method in comparison with other techniques. These metrics are defined as follows:

$$Accuracy = \frac{TP + TN}{TP + TN + FP + FN}$$

$$Sensitivity = \frac{TP}{TP + FN}$$

$$Specificity = \frac{TN}{TN + FP}$$

$$G - mean = \sqrt{Sensitivity \times Specificity}$$

$$Precision = \frac{TP}{TP + FP}$$

$$Recall = \frac{TP}{TP + FN}$$

$$F1 - score = 2\frac{Precision . Recall}{Precision + Recall}$$

Where TP, TN, FP, and FN indicate true positive, true negative, false positive and false negative, respectively.

### 5.2 Classifiers

Three different classifiers include SVM, Decision tree (C4.5) and KNN were utilized to evaluate feature selection techniques

### 6. Results

Section 6.1 and 6.2 show the results of RRQR and QR-GA, respectively.

### 6.1 RRQR results

The performance of proposed method in comparison with other techniques were shown in Figure 4, Figure 5, Figure 6, Figure 7, Figure 8, and Figure 9. The performance of feature selection methods of all datasets is shown in the radar chart and bar chart. Five evaluation metrics namely accuracy, sensitivity, specificity, F1-score, and G-mean are reported. These five evaluation criteria are portrayed in the radar charts and thereby the charts have five-sided shapes. Figure 4 shows the performances of the KNN classifier for all feature selection techniques. Figure 4(a) illustrates the radar chart of proposed technique and the other techniques in which the five metrics were computed via a KNN classifier for the top-50 features consideration. Figure 4(b) shows the same results via a bar chart. As shown in the figures, Rref is only technique which outperforms the proposed method. In terms of accuracy, the proposed method, and other techniques (except Rref) are close to each other, whereas the proposed method has significant better results in terms of sensitivity and G-mean. G-mean metric is one the most important metrics which shows the performance of classifier on imbalanced data. Consequently, it can be concluded that the performance of proposed method with respect to a KNN classifier is dependable. Figure 5 and Figure 6 depict experimental results same as Figure 4 for a SVM classifier and C4.5 classifier for 50 features, respectively. Figure 4(a), Figure 5(a), and Figure 6(a) depict symmetrical radar charts which shows the robustness and ability of the proposed method to perform well on both balanced and imbalanced data. Figure 6(b) shows that the accuracy of proposed method was highest in comparison with other techniques. Additionally, Figure 6(a) shows a symmetrical radar chart with a bigger area compared to other methods for the proposed method showing its superiority with respect to the five-evaluation metrics over other techniques.

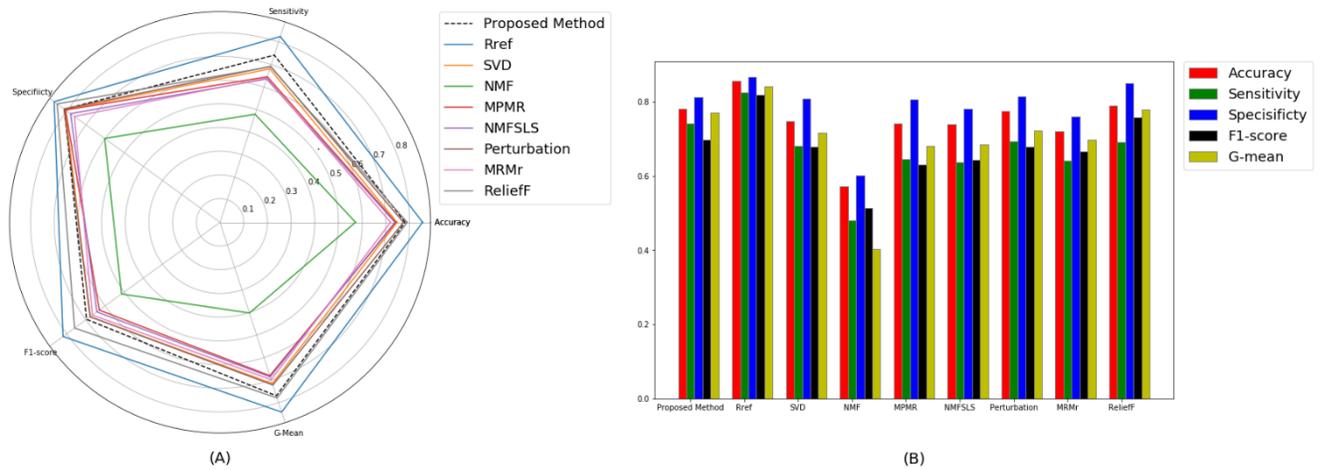

**Figure 4.** (a) Radar chart- top 50 features, (b) column chart-top 50 features of average obtained result for KNN classifier on binary datasets by considering Accuracy, Sensitivity and Specificity, F1-score and G-mean.

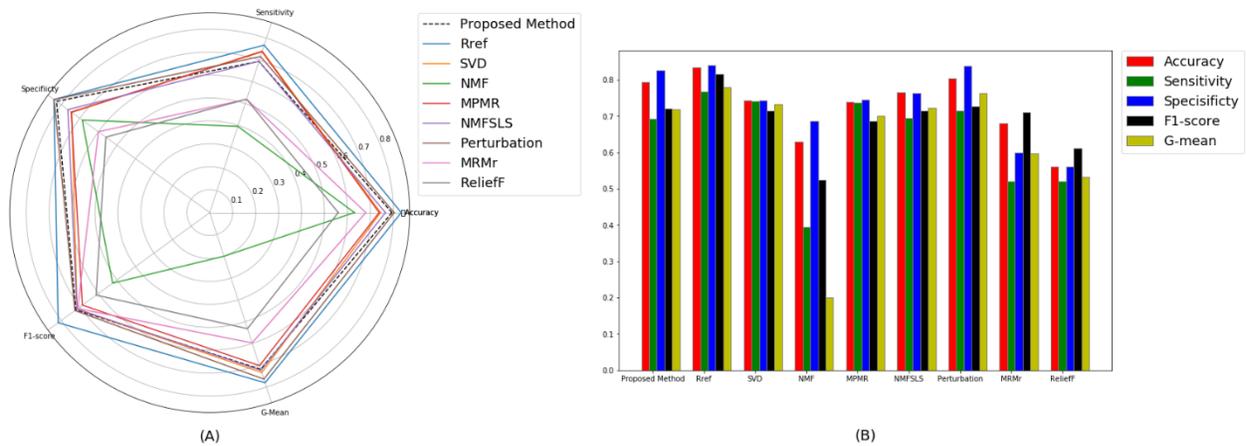

**Figure 5.** (a) Radar chart- top 50 features, (b) column chart-top 50 features of chart of average obtained result for SVM classifier on binary datasets by considering Accuracy, Sensitivity and Specificity, F1-score and G-mean.

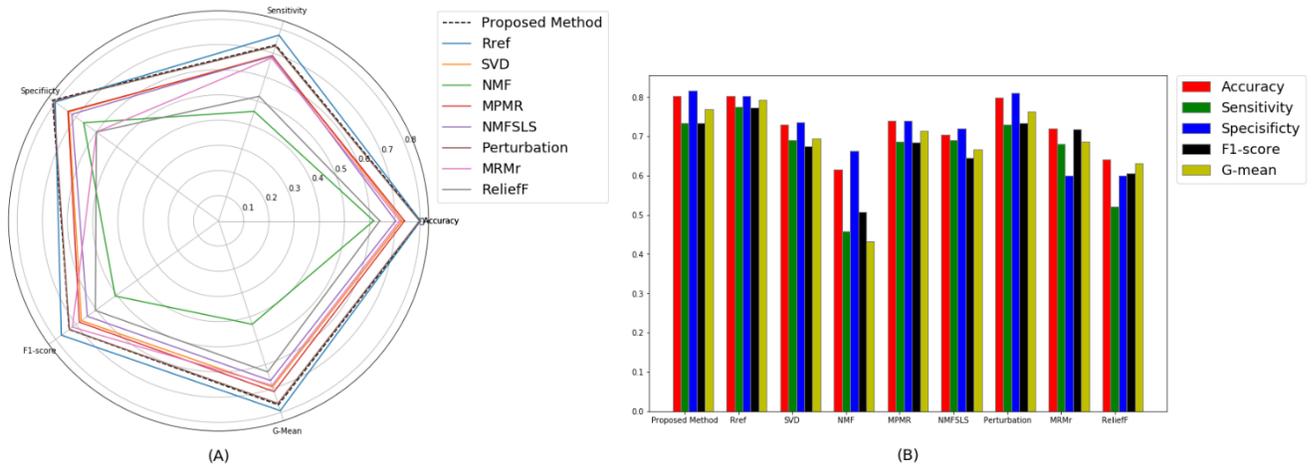

**Figure 6.** (a) Radar chart- top 50 features, (b) column chart-top 50 features of average obtained result for C4.5 classifier on binary datasets by considering Accuracy, Sensitivity and Specificity, F1-score and G-mean.

Figure 7, Figure 8 and Figure 9 depict performance of feature selection techniques for top-10 features via KNN, SVM and C4.5, respectively. Results showed that proposed method got third rank in terms of all evaluation metrics. Figure 8(a) showed that proposed method had best performance in terms of specificity evaluation metric and whereas it had weakly performance in terms of G-mean. Figure 9(a) depicted symmetrical radar chart for proposed method using C4.5.

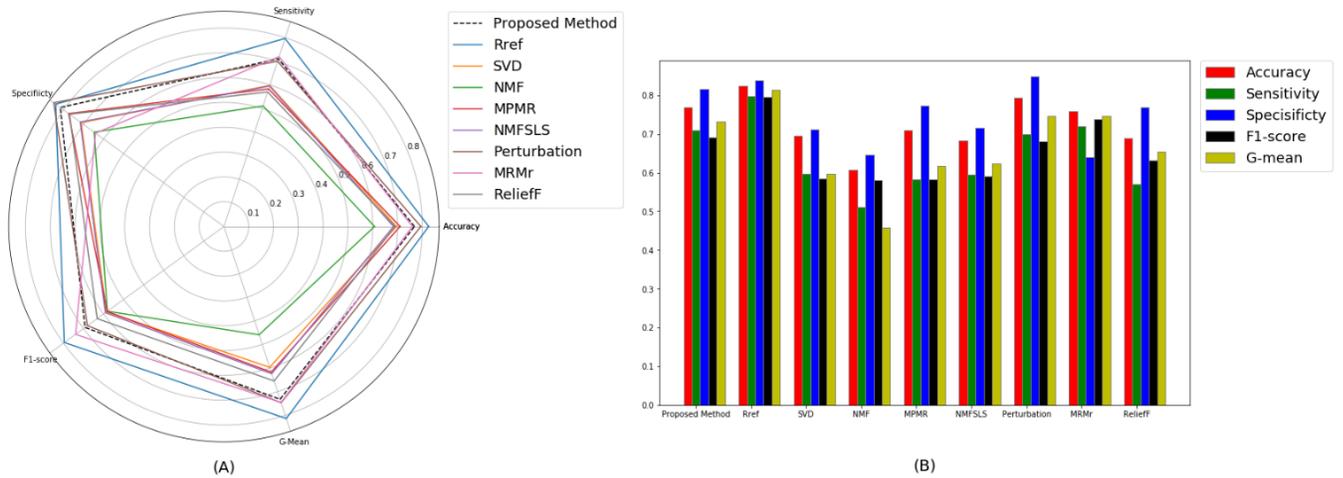

**Figure 7.** (a) Radar chart- top 10 features, (b) column chart-top 10 features of average obtained result for KNN classifier on binary datasets by considering Accuracy, Sensitivity and Specificity, F1-score and G-mean.

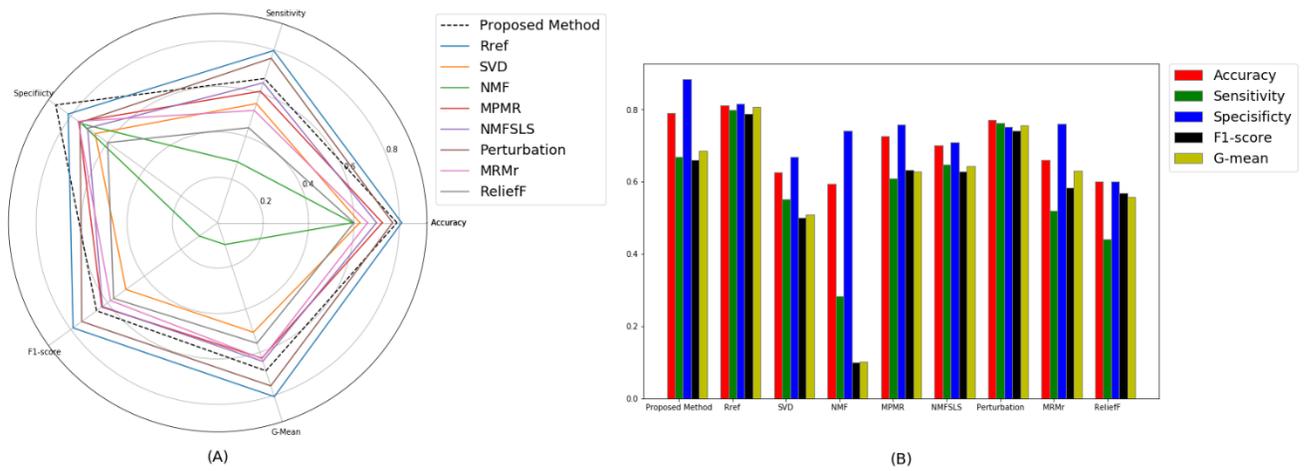

**Figure 8.** (a) Radar chart- top 10 features, (b) column chart-top 10 features of average obtained result for SVM classifier on binary datasets by considering Accuracy, Sensitivity and Specificity, F1-score and G-mean.

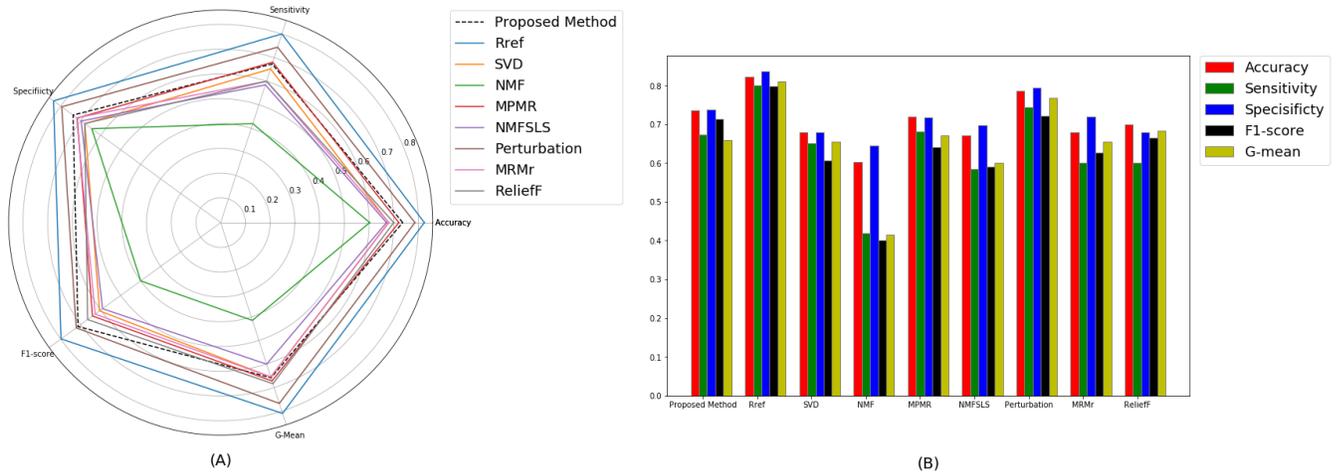

**Figure 9.** (a) Radar chart- top 10 features, (b) column chart-top 10 features of average obtained result for C4.5 classifier on binary datasets by considering Accuracy, Sensitivity and Specificity, F1-score and G-mean.

Parameter $f$ (lower bound of $R_{11}^{-1}R_{12}$) can play a significant role on improving the performance of feature selection using RRQR. Figure 10 shows the accuracy of SVM classifier for different $f$. Empirical results show that $f = [1.1\ 1.12]$ improves the performance.

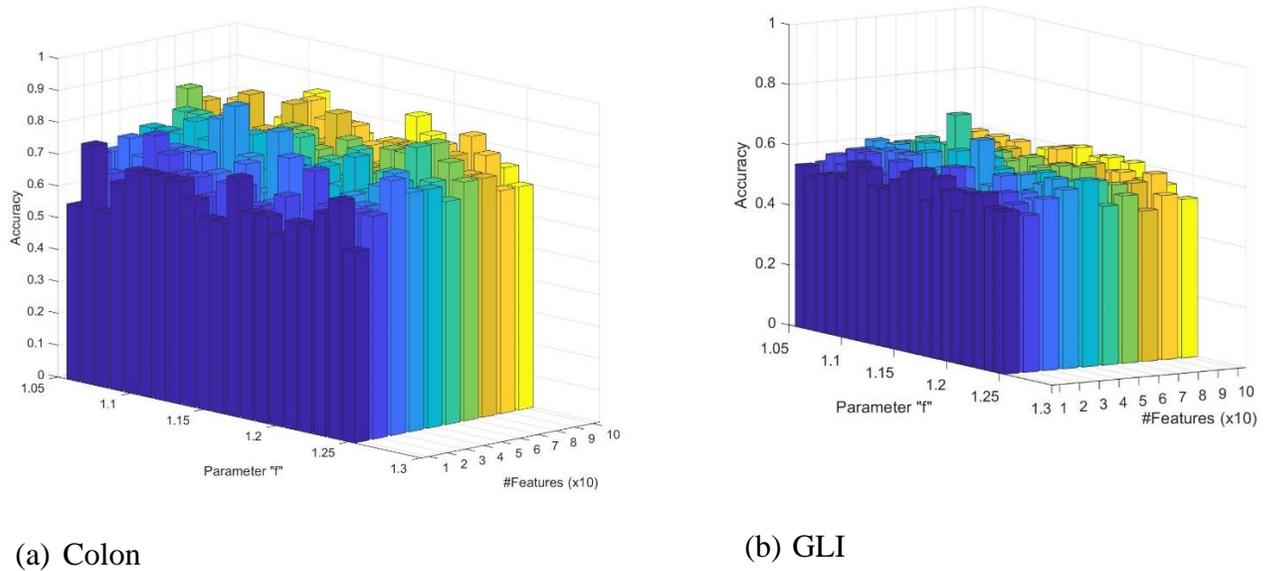

(a) Colon

(b) GLI

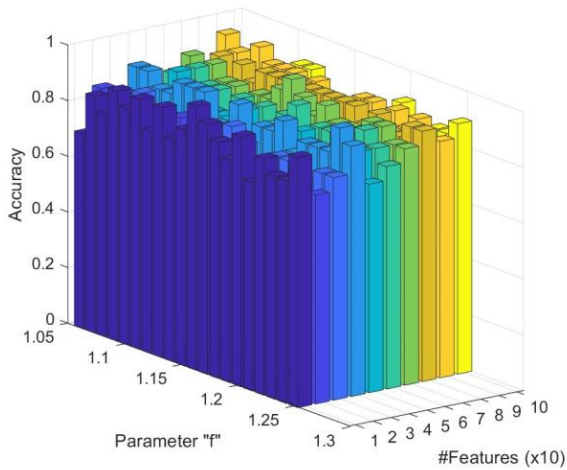
(c) Leukemia

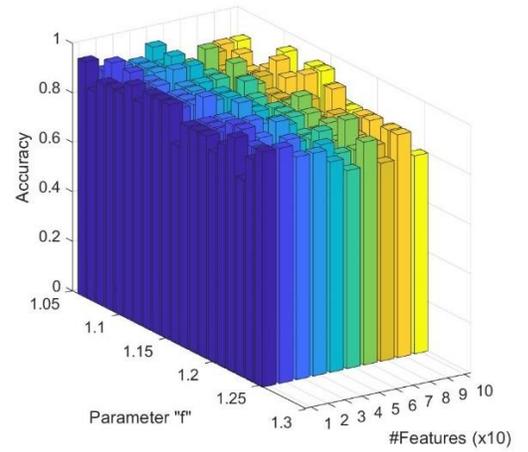
(d) Lung

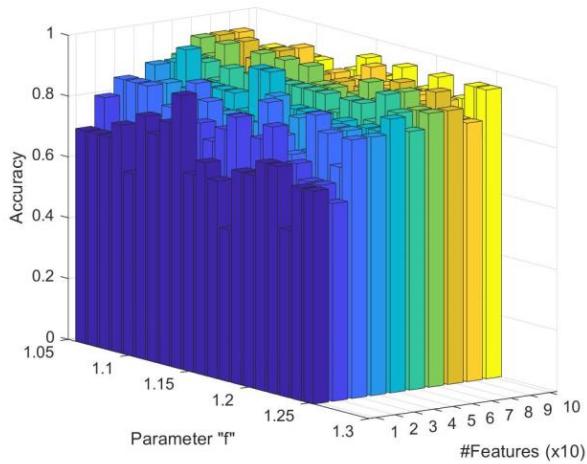
(e) Prostate

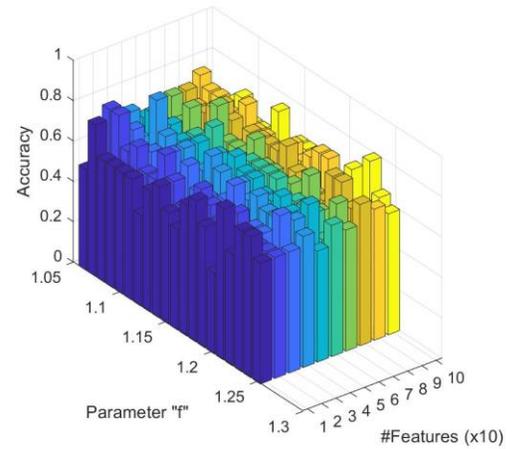
(f) SMK

**Figure 10.** Classification accuracy of Proposed method with respect to parameter "$f$" for SVM.

### 6.2 Results of NMF-QR

Table 4 and Table 5 show the performance of NMF-QR technique in comparison with RRQR with a SVM classifier used for Top-10 features and Top-50 features. The superiority of RRQR over the NMF-QR can be obviously seen. NMF-QR had better performance than RRQR for only Colon dataset with 2000 features which shows that NMF-QR can work better for lower dimension. Although Table-4 shows the NMF-QR and RRQR are close in terms of performance for SMK dataset with 19993 features, for top-50 features the RRQR had better performance than NMF-QR as shown in Table-5.

**Table 4.** Evaluation of NMF-QR technique and RRQR technique using Accuracy, F1-score and G-mean metrics for Top-10 features.

| Dataset | Accuracy % Mean | | F1-score % Mean | | G-mean % Mean | |
|---|---|---|---|---|---|---|
| | NMF-GA | RRQR | NMF-GA | RRQR | NMF-GA | RRQR |
| Ovarian | 60 | 90 | 66 | 91 | 56 | 90 |
| Colon | 72 | 70 | 59 | 54 | 69 | 64 |
| Breast | 69 | 78 | 64 | 75 | 65 | 77 |
| Prostate | 58 | 71 | 55 | 73 | 57 | 70 |
| Lung | 64 | 90 | 71 | 91 | 57 | 90 |
| SMK | 66 | 65 | 62 | 59 | 64 | 63 |
| GLI | 60 | 70 | 47 | 62 | 43 | 58 |
| Leukemia | 67 | 85 | 53 | 78 | 61 | 82 |

**Table 5.** Evaluation of NMF-QR technique and RRQR technique using Accuracy, F1-score and G-mean metrics for Top-50 features.

| Dataset | Accuracy % Mean | | F1-score % Mean | | G-mean % Mean | |
|---|---|---|---|---|---|---|
| | NMF-GA | RRQR | NMF-GA | RRQR | NMF-GA | RRQR |
| Ovarian | 80 | 97 | 78 | 98 | 79 | 98 |
| Colon | 79 | 80 | 79 | 62 | 76 | 71 |
| Breast | 74 | 82 | 71 | 80 | 73 | 79 |
| Prostate | 73 | 84 | 71 | 83 | 72 | 84 |
| Lung | 82 | 89 | 83 | 92 | 81 | 86 |
| SMK | 60 | 66 | 51 | 65 | 57 | 66 |
| GLI | 46 | 64 | 57 | 59 | 43 | 58 |
| Leukemia | 82 | 88 | 71 | 80 | 77 | 89 |

## 6.3 Results of QR-GA

Table 6 shows the performance of QR-GA technique (hybrid feature selection) when compared with RRQR with SVM used as a classifier. The results show the superiority of QR-GA over the RRQR. QR-GA can find optimum features which boost accuracy. Furthermore, hyper parameters of the SVM were tuned simultaneously with feature selection. Three evaluation metrics are reported in Table 6. In terms of metrics evaluation differences between QR-GA and RRQR were significant for Colon and Prostate datasets, and minimal for Leukemia and Lung datasets.

Table 6. Evaluation of QR-GA technique and RRQR technique using Accuracy, F1-score and G-mean metrics. Also, the optimum number of features were reported.

| Dataset | #Selected Feature | Accuracy % Mean | | F1-score % Mean | | G-mean % Mean | |
|---|---|---|---|---|---|---|---|
| | | QR-GA | RRQR | QR-GA | RRQR | QR-GA | RRQR |
| Ovarian | 96 | 98 | 85 | 98 | 86 | 98 | 84 |
| Colon | 26 | 84 | 60 | 77 | 60 | 80 | 50 |
| Breast | 56 | 88 | 83 | 86 | 81 | 85 | 82 |
| Prostate | 47 | 92 | 70 | 91 | 67 | 91 | 70 |
| SMK | 84 | 64 | 63 | 60 | 68 | 62 | 55 |
| Lung | 92 | 87 | 85 | 87 | 84 | 86 | 84 |
| GLI | 37 | 71 | 60 | 78 | 67 | 68 | 57 |
| Leukemia | 28 | 95 | 91 | 94 | 91 | 95 | 90 |

## 7. Discussion

Rank revealing QR (RRQR) matrix factorization could be used as feature selection technique by using a permutation matrix as specific ability of this matrix decomposition. It should be noted that this technique is unsupervised, whereas RREF only can only work in supervised form. In terms of parameter tuning, NMF has an $\alpha$-parameter which is a penalty parameter to keep the indicator matrix nearly orthogonal. Most studies reported that higher values of $\alpha$ can lead to more orthogonality and higher classification accuracy [35]. The number of parameters would be increased by adding regularization terms (both global and local) with these parameters having to get tunned to achieve best performance. For instance, NMFSLS has three regularization coefficients which each must individually get tuned. However, the proposed method RRQR feature selection has only one parameter, $f$ which should get tunned. The results of this study show that hybrid feature selection performs much better than RRQR. When comparing NMF-QR and RRQR, it must be considered that that RRQR is considerably faster than NMF-QR. Results showed RRQR had much better performance than NMF-QR. Filter-phase of feature selection increases the quality search of GA by removing redundant features. It is implied that redundant features are effectively constructing local optima for GA. Hyperparameters of the classifier can be tunned during feature selection using the GA in the wrapper-phase. For big-data, the computation-cost of techniques plays a significant role during method selection. The experiments of this study based on seven datasets show that RRQR is faster than NMF, NMFSLS, MRMR and SVD-feature selection and in terms of computational complexity, RRQR is less expensive than SVD and NMF based methods.

## 8. Conclusion

In this study feature selection via rank revealing-QR matrix factorization is proposed. The feature selection problem can be solved using a permutation matrix as a characteristic of the rank revealing QR matrix decomposition. Although a subspace of a matrix can be derived using SVD, it does not extract information from the matrix. The proposed method was tested on seven binary microarray datasets to evaluate its performance compared with state-of-the-art feature selection techniques. This technique can be used for different types of datasets such as business, social networks and medical. Additionally, we propose NMF-QR which is a combination of QR and NMF, and a hybrid supervised feature selection based on RRQR and a Genetic algorithm (GA). Consequently, a potential avenue for future work, would be to explore optimizing the hyperparameters of the filter-phase using an evolutionary technique such as a genetic algorithm and apply adaptive structure learning for NMF-QR to update affinity matrix in optimization process.